\documentclass[a4paper,fleqn]{cas-dc}

\usepackage[numbers]{natbib}

\usepackage{graphicx}    
\usepackage{subcaption}  

\def\tsc#1{\csdef{#1}{\textsc{\lowercase{#1}}\xspace}}
\tsc{WGM}
\tsc{QE}
\tsc{EP}
\tsc{PMS}
\tsc{BEC}
\tsc{DE}

\begin{document}
\let\WriteBookmarks\relax
\def\floatpagepagefraction{1}
\def\textpagefraction{.001}
\shorttitle{Automation in Construction}
\shortauthors{Yilong Chen et~al.}

\title [mode = title]{Learning Interaction-Aware Path Corrections from Unlabeled Trajectories: A LiDAR-Based Sim-to-Real GNN Framework for Construction Robots }

\author[1]{Yilong Chen}[type=editor,auid=000,bioid=1]
\ead{ychen3339@gatech.edu}
\cormark[1]

\credit{Writing - review \& editing, Writing - original draft, Validation, Methodology, Investigation, Formal analysis, Data curation, Conceptualization, Visualization, Software}

\affiliation[1]{organization={School of Civil and Environmental Engineering, Georgia Institute of Technology, Atlanta, GA30332},country={United States}}

\author[1]{Yong K. Cho}[]
\ead{yong.cho@ce.gatech.edu}

\credit{Conceptualization, Data curation, Investigation, Methodology, Project administration, Resources, Supervision, Writing - original draft, Writing - review \& editing}

\cortext[cor1]{Corresponding author}

\begin{abstract}
Construction robots operating near workers require navigation that can anticipate human motion while remaining robust to dynamic and unstructured site conditions. However, existing social navigation methods depend on hand-crafted interaction rules, labeled demonstrations, or robot-specific control policies, limiting their scalability and transferability in construction environments. We propose a real-time, LiDAR-only framework that learns interaction-aware path corrections from unlabeled demonstrations. Constant-velocity extrapolation generates residual learning targets, while a GNN models relationships among the robot, pedestrians, and goal context. During deployment, an extended Kalman filter tracks LiDAR detections to construct dynamic graphs, and the GNN offsets planner-generated waypoints through a planner-agnostic interface. Within bricklaying trial, the GNN reduced path irregularity by 39.2\% and increased minimum TTC. Across Dynablox sequences, it increased minimum TTC by 38.3\% while producing shorter, smoother paths than the baselines. Ablation studies supported a 1.0-s prediction interval and confirmed the benefits of target context, reflection augmentation, and feature normalization.
\end{abstract}



\begin{keywords}
Construction robots\sep 
Human-robot interaction (HRI)\sep 
Graph neural network (GNN)\sep 
Sim-to-real transfer\sep 
Dynamic object tracking\sep 
Social navigation\sep 
Self-supervised learning (SSL)
\end{keywords}

\maketitle
\section{Introduction}
Construction robot navigation has become increasingly important as autonomous and semi-autonomous mobile robots are deployed for material transport, inspection, mapping, and other repetitive operations. Unlike structured industrial facilities, construction sites evolve continuously and contain narrow, cluttered spaces shared by robots and workers performing task-driven and safety-critical activities. Worker motion in these environments is not merely random obstacle movement. It reflects ongoing tasks, spatial constraints, and interactions with other workers and equipment. A construction robot must therefore maintain navigation efficiency while anticipating potentially conflicting human motion and avoiding disruptive or unsafe behavior. Achieving this capability using limited onboard sensing and computation remains a fundamental challenge for human--robot collaboration in construction.

Conventional local planners, including sampling- and optimization-based approaches\cite{fox1997dynamic,rosmann2017integrated}, can respond effectively to geometric obstacles but generally represent pedestrians as physical or kinematic constraints. Consequently, they have limited ability to distinguish among interaction configurations that require different anticipatory responses. Social Force Model (SFM)-based methods introduce human-aware repulsion into navigation, but their behavior depends on manually specified interaction rules and parameters that may not transfer consistently across environments \cite{Chen2026Enhance}, motivating a data-driven representation of interaction behavior.

Learning-based social-navigation methods provide an alternative by learning navigation behavior from demonstrations or simulated interactions. In particular, Graph Neural Networks (GNNs) can represent a varying number of pedestrians as nodes and explicitly model their relationships through message passing. Nevertheless, existing methods commonly require manually defined social costs, labeled preferred actions, semantic scene information, or large synthetic training datasets \cite{MansoSN,bachiller2022graph}. These requirements are difficult to satisfy on construction sites, where representative interaction data are limited and social responses are rarely annotated. Moreover, end-to-end policies that directly predict control commands are often closely coupled to a particular robot or planner, limiting their integration into existing construction-robot navigation systems. A practical framework should instead learn transferable spatial corrections that can be applied through a lightweight interface to an existing local planner.

A further challenge concerns the connection between real-world perception and graph-based navigation. GNN-based interaction models require temporally consistent estimates of object position and velocity, whereas LiDAR-based dynamic-object detectors typically provide frame-wise observations whose identities can change because of occlusion, missed detections, and objects entering or leaving the sensing range. Camera-based solutions can provide additional semantic and appearance information, but their performance is affected by illumination, dust, and visual occlusion commonly encountered on construction sites. Therefore, a LiDAR-only approach capable of tracking dynamic objects and constructing interaction graphs online would improve the robustness and deployability of socially aware construction navigation.

\begin{figure*}
    \centering
    \includegraphics[width=2.0\columnwidth]{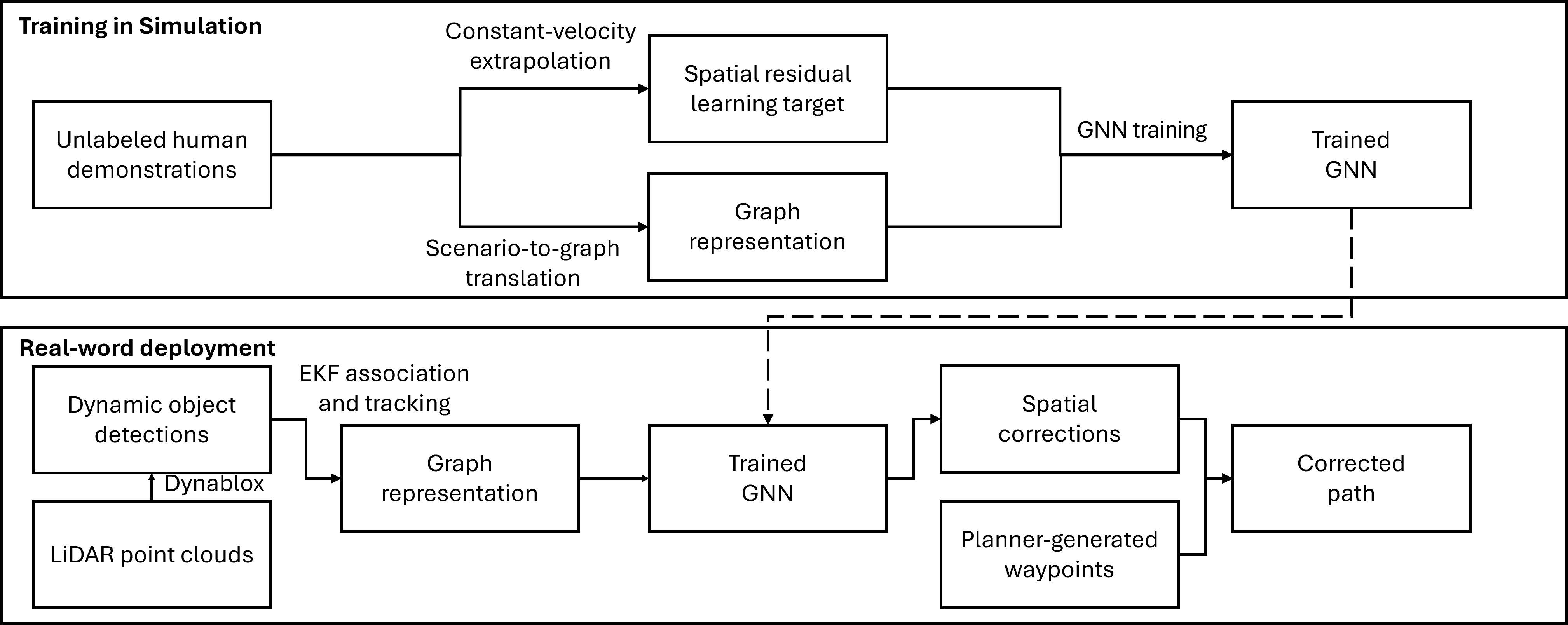}
    \caption{Overview of the proposed interaction-aware path correction framework. During training, residual targets are automatically generated from unlabeled human demonstrations using constant-velocity extrapolation, and a GNN learns relationships among the agents after scenario-to-graph translation. During deployment, Dynablox detections from LiDAR point clouds are associated and tracked using EKFs to construct dynamic graphs online. The trained GNN then predicts spatial corrections to refine planner-generated local waypoints in real time.}
    \label{fg:overall}
\end{figure*}

This study proposes a real-time, LiDAR-based framework that learns interaction-aware spatial corrections from unlabeled human demonstrations. Constant-velocity extrapolation is used to generate residual learning targets automatically, avoiding manual annotation of socially preferred trajectories. A GNN then models the relationships among the robot, nearby pedestrians, and goal context during training and predicts spatial corrections that can be applied to planner-generated paths. During deployment, frame-wise LiDAR detections are associated and tracked using an extended Kalman filter (EKF), providing the continuous positions and velocities required for online graph construction. Because the model outputs spatial residuals rather than robot-specific control commands, it provides a planner-agnostic correction interface while remaining compatible with real-time local navigation. An overview of the proposed framework is shown in Fig.~\ref{fg:overall}

\section{Related Work}
\subsection{Socially aware robot navigation}
Despite growing interest in construction robotics within both academia and industry, conventional robot local planners remain insufficient for navigation through complex dynamics in construction sites\cite{fox1997dynamic,rosmann2017integrated}. While object-aware navigation is predominantly concerned with ensuring geometric feasibility and preventing physical collisions, socially aware navigation investigates how robots can function appropriately in spaces occupied by people, with the overarching objective of achieving coexistence through awareness of human presence \cite{CHARALAMPOUS201785}, encompassing pedestrian comfort, natural human-robot interactions (HRIs), and socially appropriate robot behavior.

Human discomfort generally represents the aggravation, tension, or perception of risk induced in pedestrians by a nearby robot \cite{Gao2022}. Previous research reveals that when interacted with robots acting as passersby, or in crowded environments, people tend to prefer more conservative spatial behaviors, such as keeping a longer distance \cite{KIM2024105455}. Meanwhile, people are more likely to accept robots that exhibit human-like behavior than machine-like behavior since the former is considered more intelligent, safer, and likable \cite{kim2025understanding}. 

Both machine learning and rule-based approaches have been proposed to learn human-like behavior in construction robot navigation. Observation features from different sensor modalities are selected through paths generated by behavior cloning \cite{Chen2026AnOF}, and a Social Force Model (SFM) -enhanced Dynamic Window Approach local planner is tested on a quadruped robot navigating alongside human coworkers \cite{Chen2026Enhance}. However, imitation learning approaches may inherit the environmental distribution of their demonstrations, whereas SFM-based planners generally require scenario- and platform-dependent parameter calibration. These characteristics can limit their transfer across construction layouts, pedestrian densities, and interaction configurations.
\subsection{Graph neural networks for social navigation}
Conventional neural architectures, including Multilayer Perceptrons (MLPs) and Convolutional Neural Networks (CNNs), are generally ill-suited to graph-structured data because graphs exhibit irregular connectivity, remain invariant to permutations of node order, and may contain varying numbers of nodes and edges \cite{JIA2023104984}. Gori et al. \cite{GoriGNN} first introduced the concept of Graph Neural Networks (GNNs), in which node representations are learned through the iterative propagation of information among neighboring nodes, and subsequent research developed spatial convolution-based GNN architectures that extend convolution operations to irregular graph domains through message-passing mechanisms \cite{MicheliMP}. 

Within the Message Passing Neural Network (MPNN) framework \cite{GilmerMP}, messages are aggregated using both node attributes and edge features, allowing pairwise relationships to be explicitly represented. When applied to social navigation, this capability enables human-robot interactions to be encoded as edge attributes linking nodes that describe the states of individual agents. Manso et al. \cite{MansoSN}, for example, investigated multiple GNN architectures for assessing compliance with social-navigation principles by estimating the extent to which a robot occupying a particular position could disturb surrounding individuals. Bachiller et al. \cite{bachiller2022graph} then incorporated the historical states and motion patterns of robots and humans into a GNN to model discomfort arising from explicit interactions, movement trajectories, and velocities. 

Nevertheless, those existing approaches depend on semantically rich graph representations containing annotations of object categories, interaction engagement, and structured environmental elements, which are generally obtained from synthetic indoor datasets with predetermined layouts, constraining their applicability to construction robotics. Since construction environments rarely provide semantic annotations or pre-labeled human behaviors, their interaction patterns must instead be inferred from unprocessed spatial observations.
\subsection{Sim-to-real transfer for robot navigation}
Given the limited data available in construction, sim-to-real learning transfers models trained primarily in simulation to physical systems while mitigating discrepancies in sensing, appearance, dynamics, and environmental complexity. Common approaches include system identification, domain randomization, domain adaptation, perturbation-based robustness training, and model adaptation using limited real-world data. Although sim-to-real has been extensively studied for manipulation and grasping, comparatively few studies in major robotics surveys address mobile navigation \cite{ZhaoSim2Real}, where perception, localization, actuation, and environmental changes jointly affect transfer. Navigation therefore presents a system-level domain gap beyond visual or dynamic discrepancies alone.

Recent studies demonstrate complementary transfer strategies. A visual reinforcement learning navigation policy is trained in simulation using human interventions and demonstrations, then the visual sim-to-real gap is reduced through a denoised representation for domain adaptation \cite{WuSim2Real}. Rather than directly on simulated RGB-D images, researchers show that by operating on a domain-robust semantic-map abstraction, modular policies can transfer more reliably \cite{GervetSR}. NavRL++ systematically identifies sim-to-real gaps across sensor noise, perception failures, latency, and control-response mismatch, then introduces perturbation-aware fine-tuning and temporal reasoning for robust navigation deployment \cite{xu2026navrlsystemlevelframeworkimproving}.

Despite these advances, most sim-to-real navigation methods commonly rely on visual observations, transfer complete control policies, or require extensive system-level adaptation, making them sensitive to appearance changes, robot dynamics, and perception errors. Moreover, major studies focus on indoor or general navigation tasks rather than socially compliant navigation in dynamic, unstructured construction environments. This motivates a LiDAR-only framework that transfers domain-robust agent relationships instead of an entire navigation policy.
\subsection{LiDAR-based dynamic object tracking}
Although LiDAR Simultaneous Localization and Mapping (SLAM) is widely used in autonomous robotics, jointly moving sensors and objects complicate scan registration and static-map construction. Dynamic changes can either be short-term, such as pedestrians moving through the current field of view, or long-term, such as structural modifications to the environment \cite{slam-handbook}. For social navigation, the perception system must not only segment dynamic points but also associate them over time and estimate object-level motion states. Thus, dynamic-point segmentation and temporal tracking constitute essential steps for constructing stable pedestrian representations.

Recent studies have explored geometry- and fusion-based approaches to dynamic object detection. Dynablox \cite{Schmid_2023} integrates LiDAR observations into a TSDF map for surface reconstruction while maintaining a conservative free-space representation, where points that violate previously observed free space are identified as dynamic. Chen et al. integrate LiDAR SLAM with an occupancy grid to distinguish moving points from the registered static environment and cluster them into dynamic objects \cite{chen2025online}. A subsequent framework combines registered LiDAR point clouds with an upward-facing fisheye camera, projects LiDAR detections onto cylindrical panoramic images for semantic association, and uses these observations to update a Kalman filter for object tracking \cite{chen2026icra}.

However, none of these methods independently provide the complete LiDAR-only perception capability required by the proposed navigation framework. Dynablox robustly segments dynamic points through TSDF mapping and conservative free-space reasoning, but does not maintain persistent object identities or object-level motion states. The SLAM-occupancy-grid method similarly detects and clusters moving points without dedicated temporal association, whereas the fisheye-LiDAR framework introduces tracking but depends on an additional modality. Therefore, a LiDAR-based tracking layer is required to associate Dynablox detections across scans and continuously estimate object positions and velocities for graph construction.
\section{Methodology}
\subsection{Research Problem}
Socially aware navigation requires a mobile robot to adapt its path according to the motion of nearby workers while continuing to progress toward its navigation goal. Learning such behavior from human demonstrations is promising because demonstrations implicitly capture interaction preferences, including yielding, maintaining clearance, and returning to the original route after an encounter. However, conventional supervised approaches require manually annotated socially preferred paths or control actions. Such annotations are difficult to define consistently and expensive to obtain, particularly in construction environments containing diverse and continuously changing human-robot interactions. Therefore, the first research problem is \textit{to determine whether interaction-aware motion corrections can be learned directly from unlabeled demonstration trajectories using automatically generated training targets.}

On the other hand, transferring interaction behaviors learned in simulation to physical construction robots introduces additional challenges. Differences in sensing, coordinate systems, scene layouts, and interaction directions can cause a model to overfit to the simulated domain. Furthermore, a path-correction model coupled to a specific local planner has limited applicability across robotic platforms. Thus, the second research problem of this study is \textit{to investigate a domain-invariant graph representation in which the robot, navigation goal, and nearby moving objects are represented through robot-centric geometric and kinematic features.} The model predicts a residual spatial correction rather than planner-specific control commands, allowing the learned interaction behavior to refine nominal paths generated by different planning methods. Symmetry-aware augmentation is further investigated to improve generalization under limited and imperfect demonstration data.

A deployable system must also construct these interaction graphs from real-world sensor observations. Although LiDAR-based dynamic-object detection can identify moving regions without relying on appearance or illumination, frame-wise detections do not directly provide the temporally consistent identities and velocities required by the graph model. Accordingly, the final research problem concerns converting noisy LiDAR detections into continuous dynamic-object states suitable for online path refinement. This study addresses these interconnected problems by integrating LiDAR-based dynamic object detection and tracking with graph-based residual learning. The central research question is \textit{whether a LiDAR-only, sim-to-real GNN framework can learn planner-agnostic interaction-aware corrections from unlabeled trajectories and improve the safety, smoothness, and efficiency of construction-robot navigation in dynamic environments.}
\subsection{Training data from human demonstration}
\begin{figure}
    \centering
    \includegraphics[width=1.0\columnwidth]{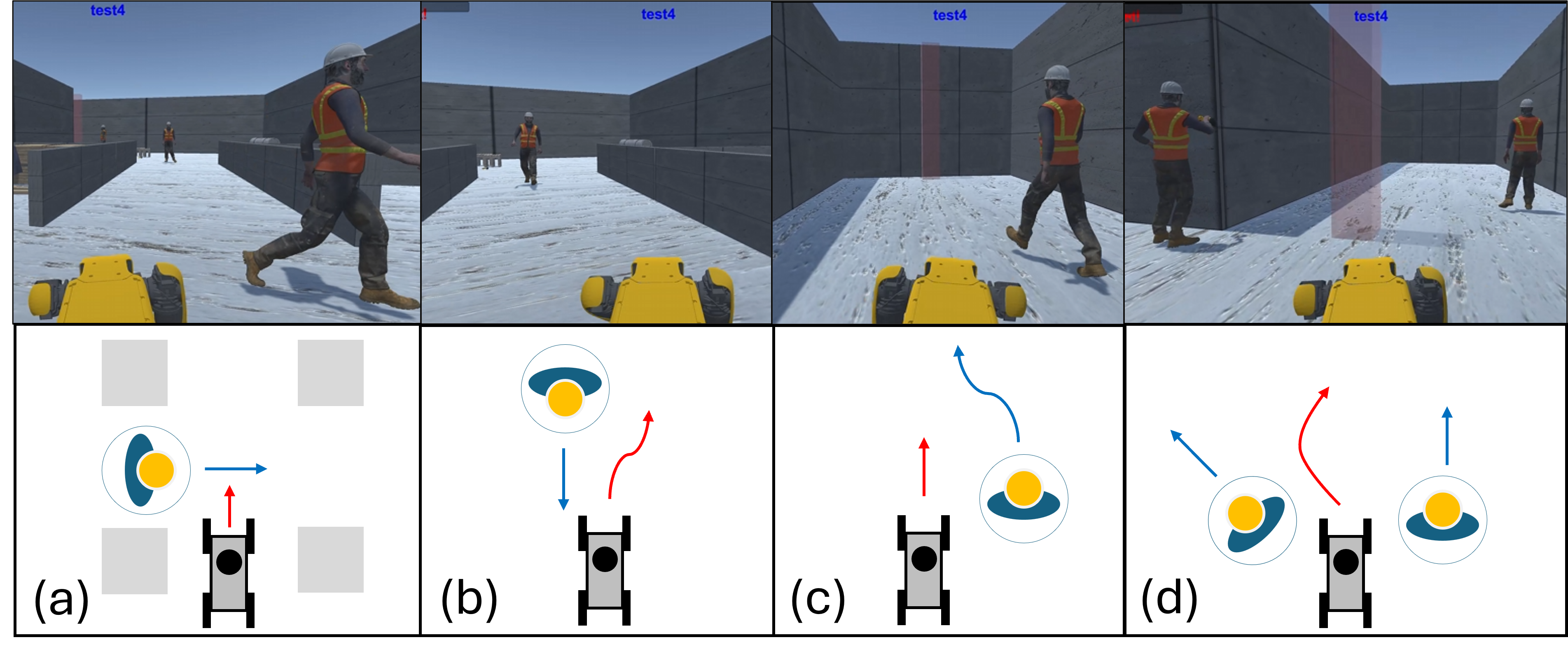}
    \caption{Simulated Environment with HRI Patterns: (a) crossing paths, (b) merging, (c) overtaking, and (d) group interaction.}
    \label{fg:hri_pattern}
\end{figure}

Our prior work \cite{kim2024much} employed Unity 3D to construct simulated construction environments for investigating HRIs under several interaction scenarios (Fig.~\ref{fg:hri_pattern}), namely path crossing, merging, overtaking, and group interaction. Within these simulations, participants operated a mobile robot using keyboard controls from a camera-based perspective and navigated it through two consecutive waypoints while preventing collisions with nearby workers. The resulting dataset in total comprised 74 trials completed by architecture, engineering, and construction professionals with relevant industry experience. For this study, the GNN model learns from 36 manually screened smooth and collision-free demonstrations using automatically generated residual targets. The recorded keyboard commands were subsequently replayed to reconstruct the corresponding simulation scenarios, from which the node and edge features at each frame were computed.
\subsection{Sim-to-real transfer learning}
\subsubsection{Domain-invariant scenario-to-graph translation}
\begin{figure}
    \centering
    \includegraphics[width=1.0\columnwidth]{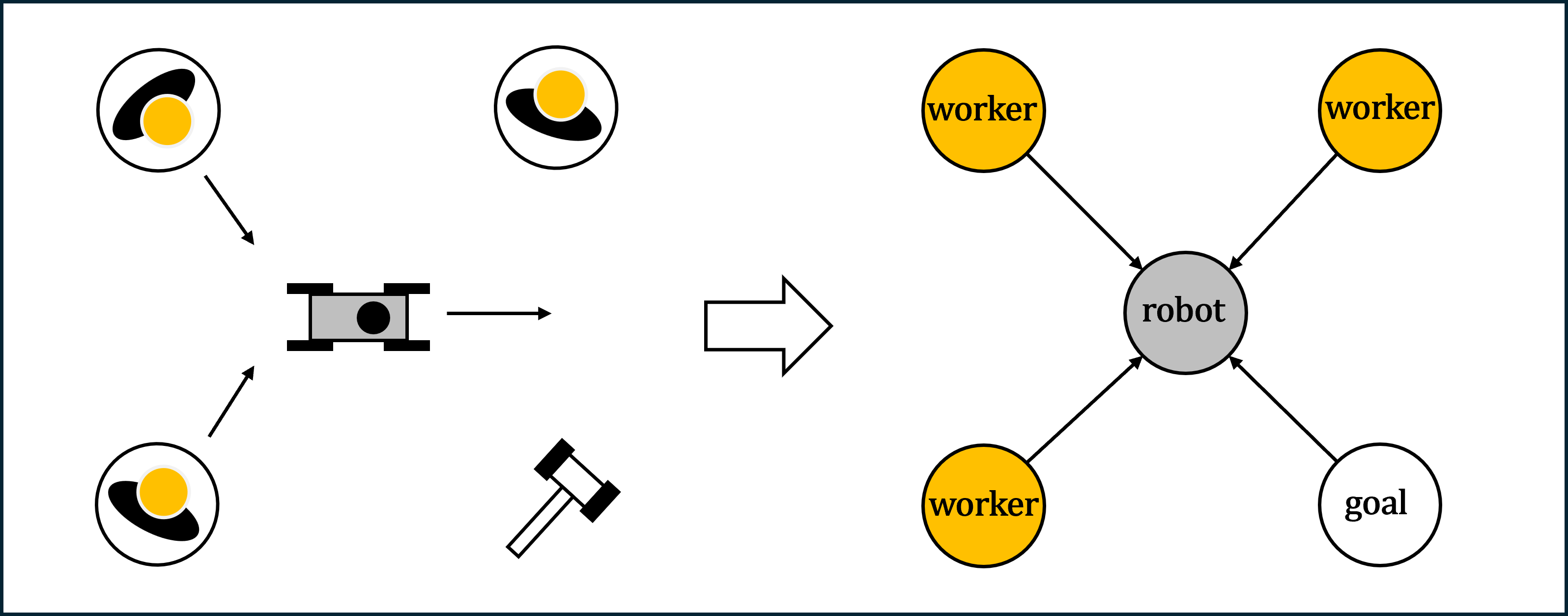}
    \caption{Scenario-to-graph translation.}
    \label{fg:scenario_to_graph}
\end{figure}

We model each scene frame as a domain-invariant graph from simulation to the real world, where agents are represented as nodes and their interactions are characterized by directed edges, as shown in Fig.~\ref{fg:scenario_to_graph} and summarized in Table ~\ref{tb:features}. The node features describe the local state of each agent within a robot-centric coordinate system and comprise its relative position $x$ and $y$, relative velocity $v_x$ and $v_y$, and a ternary indicator $r$, which is assigned a value of $1$ for the robot, $0$ for pedestrians, and $-1$ for goals. Expressing agent states relative to the robot eliminates reliance on global coordinates, thereby enhancing model generalizability across environments. During inference, the target node is omitted because it is not directly available from the LiDAR-based perception pipeline and would require additional integration with the local planner, which would reduce the planner-agnostic nature of the proposed framework.

Pairwise geometric and kinematic relationships among agents are explicitly represented through edge features. For each directed edge, these features comprise the relative displacement $x$ and $y$, relative velocity $v_x$ and $v_y$, Euclidean distance $d$, and angular direction $\sin\theta$ and $\cos\theta$ represented trigonometrically. Together, the distance and trigonometric terms capture both magnitude and orientation while preventing the discontinuities inherent in raw angular values. This separation of agent-specific states as node features and pairwise interactions as edge features enables the GNN to learn HRI patterns through message passing while accommodating variable numbers of pedestrians and diverse dynamic scene configurations. We also scaled the relative positions $x$, $y$, and velocities $v_x$, $v_y$ to the $[-1, 1]$ range, ensuring consistency with other features used during training and inference. We conducted an ablation study about normalization in the Experiment section.
\begin{table}
\caption{Node and edge features}\label{tb:features}
\begin{tabular*}{\tblwidth}{@{}LLL@{}}
\toprule
 Feature & dimension & details\\
\midrule
 Node & 5 & $[x, y, v_x, v_y, r]$ \\ 
 Edge & 7 & $[x, y, v_x, v_y, d, \sin\theta, \cos\theta]$\\
\bottomrule
\end{tabular*}
\end{table}
\subsubsection{Planner-agnostic interaction-aware path correction}
Due to the lack of explicit ground truth labels from human demonstration, we use the residual loss as the target function $L_{\text{res}}$ (Equation ~\ref{eq:l_res}), which is a summation of the error between predicted interaction-aware motion residual $G(V_{i,t}, E_{i,t})$ and constant-velocity residual $\Delta p_{i,t}$ over each trajectory $T_i$ by each participant $i$. Constant-velocity residual $\Delta p$ \cite{Chen2026Social} is used to estimate how far the robot would deviate from its current position $p_{curr}$ if it continued moving from the previous position $p_{prev}$ with the previous velocity $v_{prev}$ over a short prediction interval $\Delta t$ (Equation ~\ref{eq:delta_p}). This formulation makes spatial offset agnostic to the existing local planner.

Since constant-velocity residuals may arise from ordinary navigation behavior rather than pedestrian interaction, a heuristic filtering step was applied to retain candidate residual samples when the closest pedestrian was within \(2.5\,\mathrm{m}\) and the angle between the residual and the path direction exceeded 45°. The proximity criterion restricts the training data to locally relevant pedestrian interactions, while the angular criterion reduces predominantly longitudinal residuals associated with acceleration or deceleration. The thresholds were empirically selected to provide a balance between retaining sufficient interaction samples and excluding motion changes unlikely to represent lateral interaction responses.

\begin{figure}
    \centering
    \includegraphics[width=1.0\columnwidth]{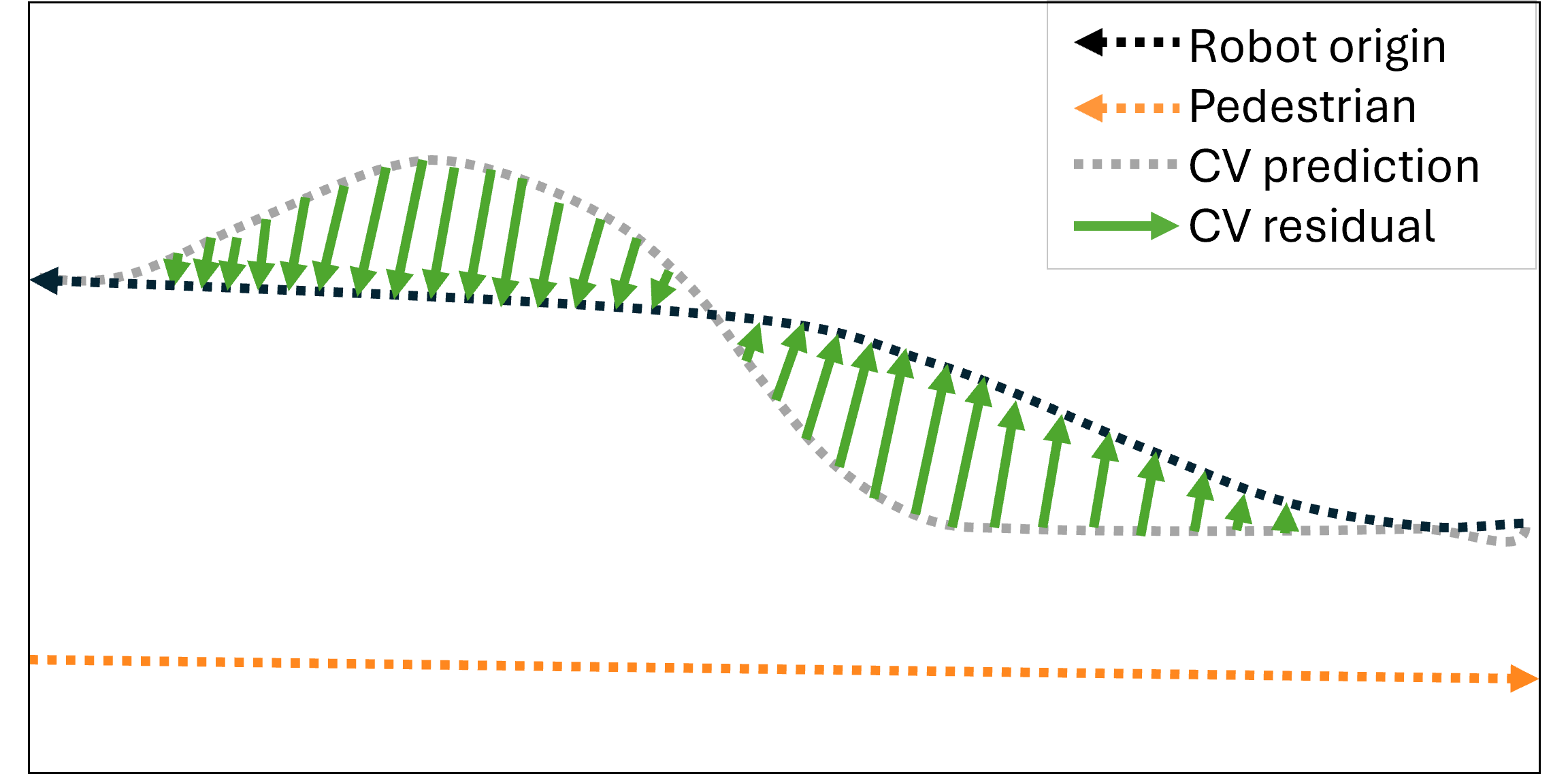}
    \caption{Constant-velocity residual.}
    \label{fg:cv_residual}
\end{figure}

Fig.~\ref{fg:cv_residual} illustrates an example about how the constant-velocity residual influences the robot's trajectory. When a robot encounters a pedestrian in a crowded environment, it tends to first yield and maintain a greater distance. Once it has passed the pedestrian, it resumes its original route. Additionally, as the robot approaches its target node, it exhibits a significant deviation relative to its position at an earlier time point. In practice, we set the prediction interval to \(1.0\,\mathrm{s}\), and the total number of frames is therefore 28622. We also conducted an experiment to investigate the effect of different values of prediction interval on the results.

\begin{align}
L_{\text{res}} &= \sum_{i=1}^{N} \sum_{t=1}^{T_i} \left\| G(V_{i,t}, E_{i,t}) - \Delta p_{i,t} \right\|^2 \label{eq:l_res} \\
V^{'}_{i,t} &= V_{i,t}\text{diag}(1, -1, 1, -1, 1)\label{eq:v_flip}\\
\Delta p &= p_{\text{curr}} - (p_{\text{prev}} + v_{\text{prev}} \Delta t) \label{eq:delta_p}
\end{align}

\subsubsection{Symmetry-aware GNN training}
Although message-passing GNNs are inherently equivariant to permutations of node ordering, they are not generally equivariant to geometric reflections when Cartesian node and edge features are processed by unconstrained neural networks \cite{wang2023graphneuralnetworklocal}. To increase the generalizability of the model across geometrically symmetric layouts, we conduct data augmentation by flipping the node features $V_{i,t}$ along x-axis (Equation ~\ref{eq:v_flip}) and get node features $V'_{i,t}$ and corresponding edge features of augmented data. We investigate the influence of data augmentation in the experiment section.

During training, since the graph structure is relatively simple and the GNN can easily overfit to the training data, we select a smaller hidden dimension of $64$ and larger batch size at $1024$. The GNN employs a shared node encoder and edge encoder followed by multiple message-passing layers with drop rate of $0.1$, and outputs a 2-dimensional vector corresponding to the robot's interaction-aware path correction. We train the model using Huber loss to regress the residual offsets and optimize with the Adam optimizer at a learning rate initialized at $0.0005$.
\subsection{LiDAR-based dynamic object tracking}
\begin{figure}
    \centering
    \includegraphics[width=1.0\columnwidth]{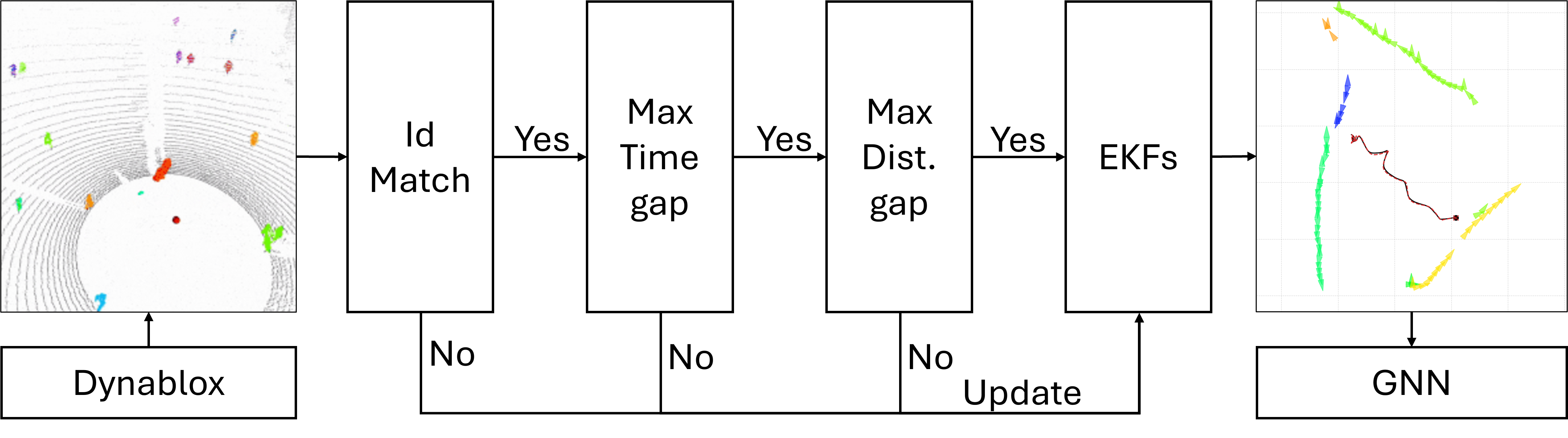}
    \caption{Dynablox-EKF-GNN pipeline.}
    \label{fg:dynablox_ekf_gnn}
\end{figure}

While Dynablox \cite{Schmid_2023} provides LiDAR-based dynamic object detection results, because of its conservative estimation of free space, noise in the real word, and absence of velocity estimation, the segment result cannot be directly applied to construct the graph node and edge features. We propose a LiDAR-based dynamic object tracking pipeline combined with Dynablox and Extended Kalman Filter (EKF) as shown in Fig.~\ref{fg:dynablox_ekf_gnn}. For each dynamic object detected, we first match the Id in the existing EKF dictionary, and then verify if it is within the maximum distance gap and maximum time gap with the detected object with the same Id. If all the requirements are met, the corresponding EKF are predicted and updated based on the current observation. Otherwise, a new EKF will be created and the old one will be deleted. This pipeline will enable the robot to track continuous moving objects. And the detected objects will be used with the robot node to construct graph for feeding into the GNN model. In practice, the maximum distance gap and maximum time gap is set to \(0.5\,\mathrm{m}\) and \(0.2\,\mathrm{s}\), respectively.

\section{Experiments}
\subsection{Datasets}
\subsubsection{Bricklaying trial}
\begin{figure}
    \centering
    \includegraphics[width=1.0\columnwidth]{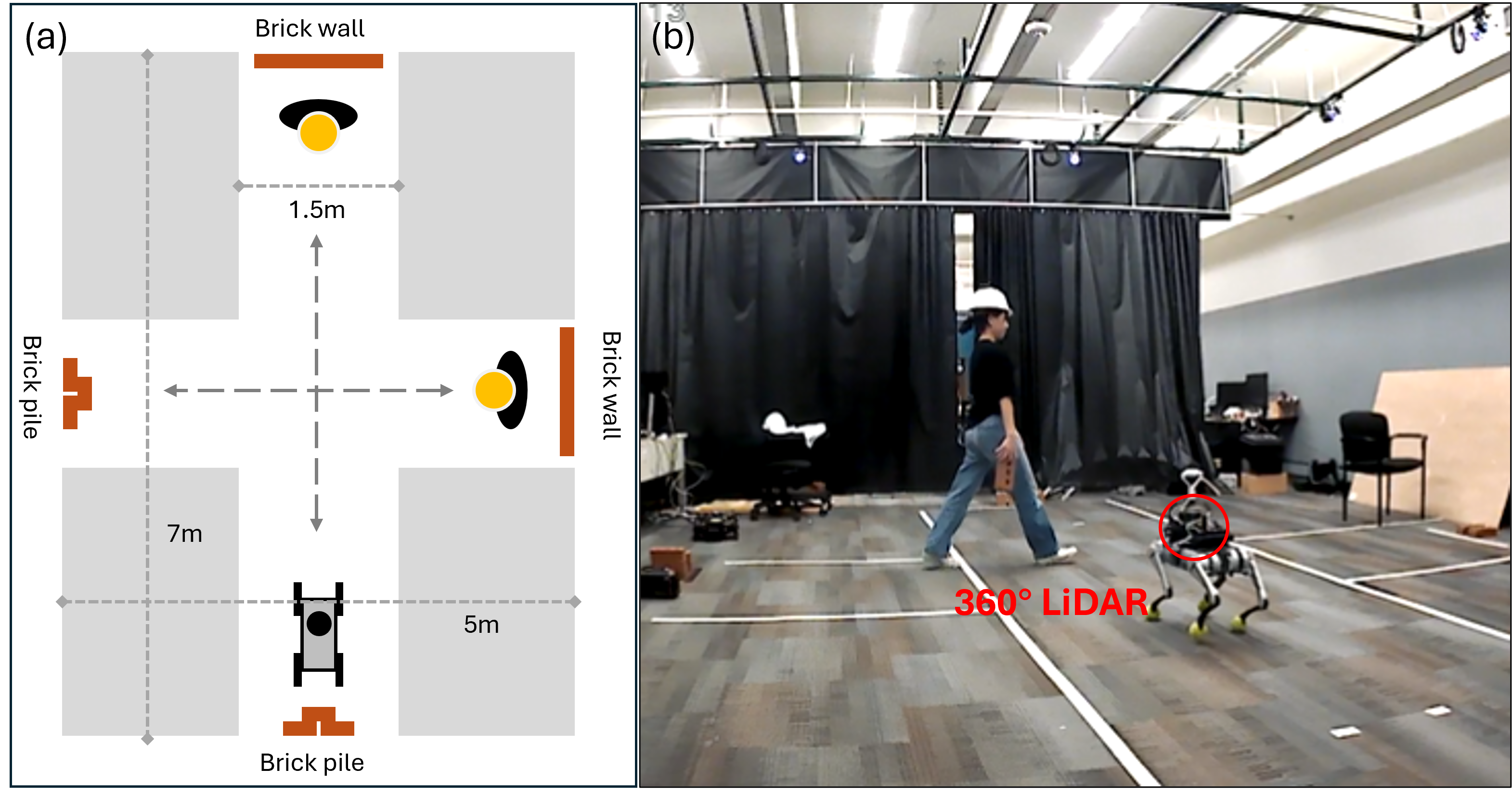}
    \caption{Bricklaying Experiment: (a) testbed layout and (b) quadruped robot with LiDAR sensor.}
    \label{fg:bricklaying}
\end{figure}

To assess the performance of the proposed GNN local planner in construction environments, an indoor bricklaying testbed was established based on the layouts adopted by \cite{kim2025understanding}, as illustrated in Fig.~\ref{fg:bricklaying}. The testbed occupied a \(5\times7\,\mathrm{m}\) rectangular area and featured a cross-shaped corridor with a width of \(1.5\,\mathrm{m}\), where brick piles were positioned at both ends of each corridor branch. While the quadruped robot with Velodyne VLP-16 LiDAR sensor repeatedly navigated along the longer corridor, two workers transported 10 bricks between its two ends and progressively constructed a wall. Apart from being instructed to avoid collisions, the workers received no predefined directions regarding when to cross or stop, allowing the cross-shaped configuration to naturally generate the HRI patterns depicted in Fig.~\ref{fg:hri_pattern}. The positions and velocities of dynamic objects were estimated using the aforementioned LiDAR-based pipeline in the Methodology section, and these estimates were subsequently used to construct the node and edge features supplied to the GNN. Its raw LiDAR measurements allow the robustness of the complete Dynablox-EKF-GNN pipeline to be evaluated without camera-derived observations, in realistic perception uncertainty. Executed on the robot's onboard CPU, the object-detection module operated at \(10\,\mathrm{Hz}\), corresponding to the control frequency of the local planner.
\subsubsection{Dynablox dataset}
\begin{figure}
    \centering
    \includegraphics[width=1.0\columnwidth]{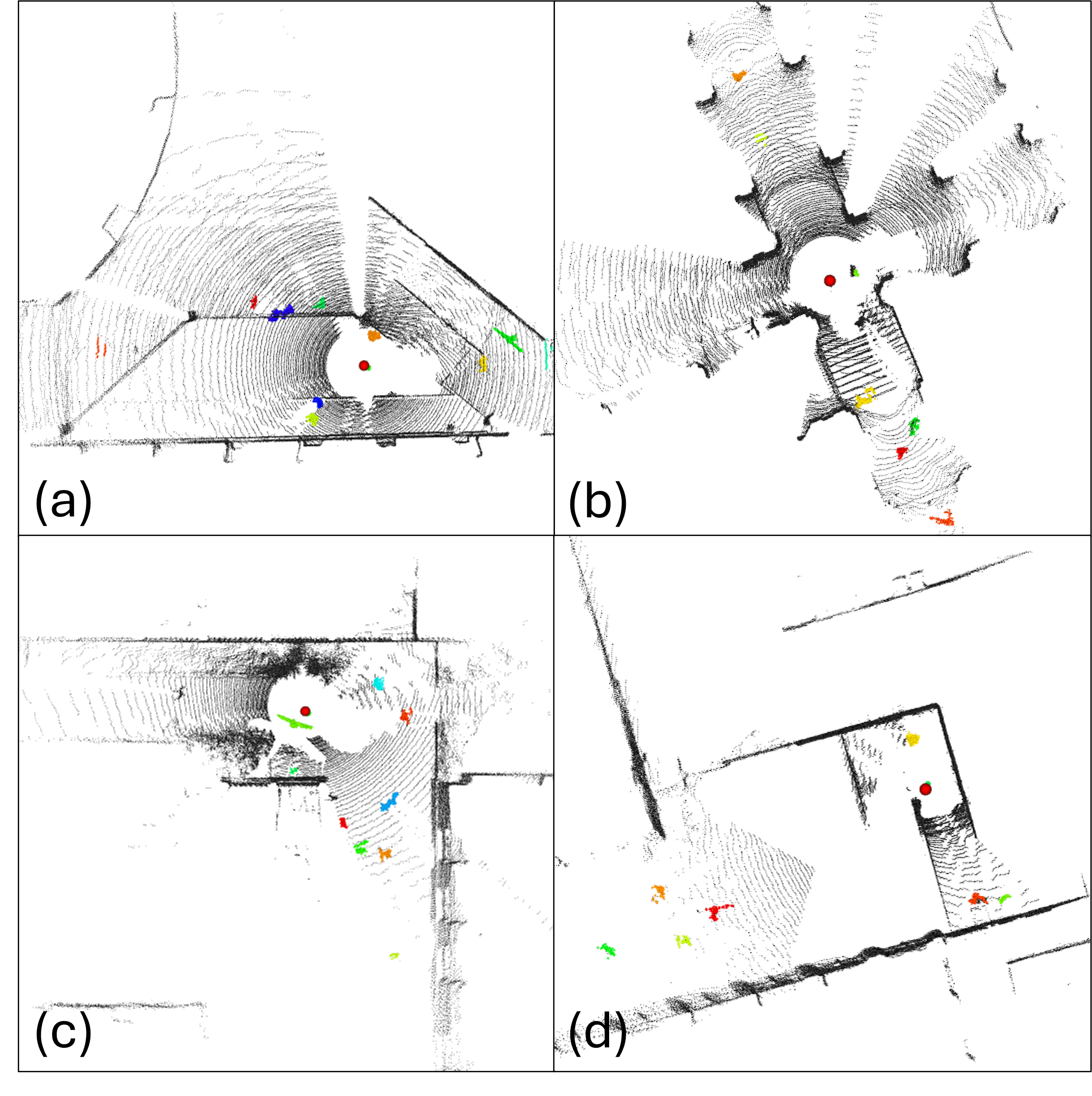}
    \caption{Dynablox dataset: (a) corridor, (b) hg, (c) ramp, and (d) stairs.}
    \label{fg:dynablox_dataset}
\end{figure}

The Dynablox dataset \cite{Schmid_2023} is a public real-world dataset developed for evaluating LiDAR-based dynamic object detection in complex environments. It comprises eight sequences recorded using a 128-beam Ouster OS0 LiDAR and covers diverse indoor and outdoor settings, including corridors, ramps, stairs, cluttered spaces, and multilevel structures (Fig.~\ref{fg:dynablox_dataset}). The sequences contain more pedestrians and atypical moving objects under challenging conditions such as occlusion and irregular motion. We select this dataset as an additional test set because it differs substantially from our structured bricklaying experiment in sensor platform, environment geometry, number of objects and object motion. While the bricklaying dataset evaluates the planner under construction-specific HRIs, the Dynablox sequences provide an independent, publicly available benchmark for examining whether the learned corrections transfer to previously unseen real-world environments and trajectories. Testing on this dataset therefore complements the task-specific experiments by assessing cross-domain generalization. Unlike the real-time, closed-loop bricklaying experiment, the proposed framework was evaluated through real-time-rate sequential processing of recorded LiDAR data. Pedestrian motions reflect responses to the original planner rather than to the corrected trajectories. In practice, we tested seven data sequences including corridor1, corridor2, hg1, hg2, ramp1, ramp2, and stairs, as shown in Table ~\ref{tb:main_result}. The indoor dataset was removed due to the short duration.
\subsection{Metrics}
Robot navigation performance is evaluated across four complementary dimensions: efficiency, smoothness, physical safety, and social compliance, as summarized in Table ~\ref{tb:metrics}. Navigation efficiency is quantified by path length because shorter trajectories generally require less travel time and energy when the start and goal conditions are fixed. Trajectory smoothness, or motion naturalness, is assessed using path irregularity \cite{GuzziHuman}, which measures unnecessary rotation relative to the distance traveled with lower values indicating a more direct and less oscillatory trajectory. Physical safety is evaluated using the minimum robot-pedestrian distance, as greater separation reduces the likelihood of collision and uncomfortable close encounters. Finally, social compliance is measured using the minimum time-to-collision (TTC) \cite{BiswasSocNav}, computed for each robot-pedestrian pair by extrapolating their trajectories under a constant-velocity assumption \cite{Francis2025}. TTC captures the temporal urgency of an interaction and distinguishes converging motions from geometrically close but nonconflicting movements. Together, these four metrics provide complementary measures of whether a planner produces short, smooth, spatially safe, and anticipatory socially compliant trajectories.
\begin{table}
\caption{Evaluation metrics}\label{tb:metrics}
\begin{tabular*}{\tblwidth}{@{}LLL@{}}
\toprule
 Dimension & Metric & Unit\\
\midrule
 Efficiency & path length & $meter$ \\ 
 Smoothness & path irregularity & $radian/meter$\\
 Physical safety & closest pedestrian distance & $meter$\\
 Social compliance & minimum time-to-collision & $second$\\
\bottomrule
\end{tabular*}
\end{table}
\subsection{Baselines}
\subsubsection{SFM-induced spatial offset}
In our earlier study \cite{Chen2026Enhance}, the repulsive forces \cite{Helbing_1995} induced by pedestrians, $F_i(t)$, were applied as offsets to transform the original path waypoint $p_i(t)$ into an adjusted waypoint $p^{o}_{i}$, as formulated in Equation~\ref{eq:sfm_offset}. Pedestrian locations were projected over an interception interval $\Delta t$ as $r_\beta + v_\beta \Delta t$, rather than being represented solely by their current positions $r_\beta$ (Equation~\ref{eq:sfm_force}). Incorporating these predicted positions enables the robot to anticipate pedestrian motion and consequently exhibit more socially responsive navigation behavior. Nevertheless, the SFM-enhanced DWA planner remains limited in its capacity to comprehensively capture complex and context-dependent HRIs at scale because the SFM depends on manually calibrated parameters whose performance is often sensitive to variations in environmental conditions. In this research, we use SFM-induced planner as a baseline for evaluating our proposed GNN-based planner. Note that in the previous study we only offset the waypoints generated by the DWA planner, but will predict for every frame in this paper for a fair comparison.
\begin{align}
p^{o}_{i}(t) &= p_i(t) + F_i(t) \label{eq:sfm_offset} \\
F_i(t) &= \sum F_{i\beta}(e_i, r_i - (r_\beta + v_\beta \Delta t)) \label{eq:sfm_force}
\end{align}
\subsubsection{MLP-based interaction pool net}
We select the MLP-based interaction pool network as a strong non-message-passing baseline to determine whether the performance gains of our GNN arise from explicit graph reasoning rather than simply from learning nonlinear functions of pairwise interaction features. The pool net encodes each robot-to-agent edge using a shared MLP and aggregates the resulting embeddings through permutation-invariant, attention-weighted maximum pooling, supplemented by an interaction-count feature. Thus, it naturally accommodates a variable number of surrounding agents and remains insensitive to their ordering while retaining salient geometric and kinematic information from the edge features. However, unlike GNN, it does not encode node states or iteratively propagate information through the interaction graph. Consequently, it may not be able to explicitly model higher-order dependencies among the robot, goal, and nearby agents. Its combination of flexible nonlinear encoding, attention-based aggregation, maximum pooling, and crowd-size information nevertheless provides a more competitive baseline than other neural networks. Comparing the two models therefore isolates the value of structured node representations and multi-layer relational message passing for learning interaction-aware path corrections.
\subsection{Ablation study}
\subsubsection{Ablation on different prediction interval}
The prediction interval $\Delta t$ directly determines the constant-velocity residual used as the learning target and therefore affects both its physical meaning and magnitude. A short interval produces small residuals that may be dominated by trajectory discretization, sensor noise, or minor control fluctuations, whereas a long interval can generate larger corrections that combine social responses with acceleration, turning, goal approaching, and accumulated prediction error. Consequently, the selected $\Delta t$ controls the temporal scale of the behavior learned by the GNN and may influence prediction accuracy, path smoothness, safety, and the number of valid training samples. An ablation study is therefore necessary to verify that the selected value captures meaningful interaction responses without introducing excessive motion-model error. 
\subsubsection{Ablation on different model variant}
An ablation study is necessary to quantify the contribution of each design component and determine whether the model's performance results from meaningful interaction learning. Removing the target node evaluates whether explicit goal information enables the GNN to distinguish socially motivated deviations from ordinary goal-directed motion. Disabling data augmentation assesses whether mirrored interaction scenarios improve generalization across spatially symmetric layouts that are not sufficiently represented in demonstrations. Finally, disabling feature normalization evaluates whether scaling heterogeneous position, velocity, and distance features to comparable ranges improves optimization stability and prevents large-magnitude variables from dominating the learned representation. 
\begin{table*}[width=1.8\linewidth,cols=11,pos=h]
  \setlength{\tabcolsep}{2.4pt}
  \caption{Evaluation results on the Dynablox dataset and bricklaying trial. "Dynablox avg." reports the mean across seven sequences. Bold indicates the best result. Standard deviations are calculated within each sequence and across sequences for "Dynablox avg.".}\label{tb:main_result}
\begin{tabular*}{\textwidth}{
  >{\raggedright\arraybackslash}p{1.5cm} 
  >{\raggedright\arraybackslash}p{1.0cm} 
  ccccccccc @{}}
\toprule
\multirow{2}{*}{Measure} & \multirow{2}{*}{Traj.} & \multicolumn{9}{c}{Dataset}\\
\cmidrule(l){3-11}
 & & corridor1 & corridor2 & hg1 & hg2 & ramp1 & ramp2 & stairs & Dynablox avg. & bricklaying\\
\midrule
\multirow{4}{*}{$\begin{array}{l} \text{Path} \\ \text{length} \\(m)\downarrow\end{array}$} & Origin & 33.67 & 33.82  & 32.68 & 34.61 & 51.89 & 31.95 & 105.15 & 46.25$\pm$24.89 & 58.96\\
\cmidrule(l){2-11}
 & GNN & 40.72 & 35.24  & 31.30 & 35.06 & 53.99 & 33.99 & 113.92 & \textbf{49.17$\pm$27.33} & \textbf{63.25}\\
 & SFM & 35.26 & 37.13  & 40.91 & 36.61 & 62.46 & 34.16 & 114.85 & 51.63$\pm$27.35 & 63.97\\
 & MLP & 36.39 & 42.10  & 36.90 & 41.64 & 62.36 & 54.37 & 122.65 & 56.63$\pm$28.36 & 82.65\\
 \midrule
\multirow{4}{*}{$\begin{array}{l} \text{Path} \\ \text{irregularity} \\(rad/m)\downarrow\end{array}$} & Origin & 0.26 & 0.29  & 0.61 & 0.47 & 0.22 & 0.24 & 1.79 & 0.55$\pm$0.52 & 7.22\\
\cmidrule(l){2-11}
 & GNN & 1.56 & 1.57  & 1.32 & 1.06 & 1.29 & 1.09 & 1.84 & \textbf{1.39$\pm$0.26} & \textbf{4.39}\\
 & SFM & 1.06 & 1.62  & 2.74 & 1.32 & 2.05 & 1.28 & 3.44 & 1.93$\pm$0.81 & 7.87\\
 & MLP & 1.29 & 2.02  & 1.63 & 2.03 & 1.86 & 1.39 & 2.56 & 1.83$\pm$0.40 & 5.28\\
\midrule
\multirow{4}{*}{$\begin{array}{l} \text{Closest} \\ \text{pedestrian} \\ \text{distance} \\(m)\uparrow\end{array}$} & Origin & 1.68$\pm$0.61 & 1.75$\pm$0.49  & 1.82$\pm$0.54 & 2.06$\pm$0.38 & 1.71$\pm$0.44 & 1.59$\pm$0.48 & 1.84$\pm$0.52 & 1.78$\pm$0.49 & 1.56$\pm$0.49\\
\cmidrule(l){2-11}
 & GNN & 1.76$\pm$0.62 & 1.77$\pm$0.49  & 1.89$\pm$0.53 & 2.07$\pm$0.38 & 1.74$\pm$0.44 & 1.66$\pm$0.49 & 1.85$\pm$0.52 & 1.82$\pm$0.49 & 1.58$\pm$0.50\\
 & SFM & 1.73$\pm$0.57 & 1.78$\pm$0.46  & 1.86$\pm$0.51 & 2.09$\pm$0.37 & 1.75$\pm$0.43 & 1.63$\pm$0.48 & 1.88$\pm$0.49 & 1.82$\pm$0.47 & 1.57$\pm$0.49\\
 & MLP & 1.80$\pm$0.75 & 1.88$\pm$0.58  & 1.81$\pm$0.55 & 1.94$\pm$0.51 & 1.77$\pm$0.57 & 1.71$\pm$0.68 & 1.89$\pm$0.53 & \textbf{1.83$\pm$0.61} & \textbf{1.61$\pm$0.60}\\
\midrule
\multirow{4}{*}{$\begin{array}{l} \text{Minimum} \\ \text{TTC} \\(s)\uparrow\end{array}$} & Origin & 0.19$\pm$0.44 & 0.46$\pm$0.02  & 1.68$\pm$3.73 & 3.83$\pm$4.17 & 0.27$\pm$0.41 & 0.38$\pm$0.65 & 2.17$\pm$3.98 & 1.28$\pm$1.92 & 0.44$\pm$1.17\\
\cmidrule(l){2-11}
 & GNN & 0.32$\pm$0.83 & 0.45$\pm$1.01  & 2.09$\pm$4.07 & 6.48$\pm$11.86 & 0.43$\pm$0.55 & 0.42$\pm$0.66 & 2.24$\pm$3.93 & \textbf{1.77$\pm$3.27} & \textbf{0.51$\pm$1.37}\\
 & SFM & 0.20$\pm$0.46 & 0.69$\pm$1.32  & 1.54$\pm$3.42 & 4.45$\pm$4.43 & 0.56$\pm$1.20 & 0.41$\pm$0.67 & 2.49$\pm$4.29 & 1.48$\pm$2.25 & 0.50$\pm$1.36\\
 & MLP & 0.20$\pm$0.52 & 0.34$\pm$0.52  & 1.84$\pm$4.09 & 4.01$\pm$13.83 & 0.23$\pm$0.55 & 0.25$\pm$0.55 & 2.04$\pm$2.69 & 1.27$\pm$3.25 & 0.42$\pm$1.07\\
\bottomrule
\end{tabular*}
\end{table*}
\section{Results}
\subsection{Bricklaying trial results}
Table~\ref{tb:main_result} summarizes the evaluation results for the bricklaying trial. Compared with the original path, the GNN increased the path length from \(58.96\) to \(63.25\,\mathrm{m}\) (7.3\%), indicating that additional travel was required to respond to nearby workers. Nevertheless, it reduced path irregularity from \(7.22\) to \(4.39\,\mathrm{rad/m}\) (39.2\%), demonstrating that the corrected trajectory was substantially smoother than the original path. The GNN also maintained a same level of closest pedestrian distance (\(1.58\,\mathrm{m}\)) as the Origin (\(1.56\,\mathrm{m}\)), and increased the minimum TTC from \(0.44\) to \(0.51\,\mathrm{s}\). These results suggest that the learned correction improved trajectory smoothness and anticipatory interaction safety while incurring a moderate increase in travel distance.

Compared with the SFM and MLP baselines, the GNN generated the shortest path and achieved the lowest path irregularity. Its path was \(0.72\,\mathrm{m}\) shorter than that of the SFM and \(19.40\,\mathrm{m}\) shorter than that of the MLP, while its irregularity was \(44.2\%\) and \(16.9\%\) lower, respectively. The GNN also produced a same degree of minimum TTC (\(0.51\,\mathrm{s}\)) as the SFM (\(0.50\,\mathrm{s}\)), and more clearly outperformed the MLP (\(0.42\,\mathrm{s}\)). Although the MLP achieved the greatest closest pedestrian distance (\(1.61\,\mathrm{m}\)), it generated a considerably longer path and a lower TTC. Overall, the GNN provided the most favorable balance between efficiency, smoothness, pedestrian clearance, and anticipatory collision avoidance in the construction-specific bricklaying trial.
\subsection{Dynablox dataset results}
Across the seven Dynablox sequences, the GNN increased the average path length from \(46.25\) to \(49.17\,\mathrm{m}\), corresponding to a 6.3\% increase over the original trajectories. Path irregularity also increased from \(0.55\) to \(1.39\,\mathrm{rad/m}\), showing that interaction-aware corrections introduced additional turning into the originally direct trajectories. In return, the closest pedestrian distance increased from \(1.78\) to \(1.82\,\mathrm{m}\), while the minimum TTC increased from \(1.28\) to \(1.77\,\mathrm{s}\), an improvement of 38.3\%. Thus, the GNN traded a modest reduction in trajectory efficiency and smoothness for greater spatial clearance and a substantially larger temporal safety margin.

Among the three trajectory-refinement approaches, the GNN produced the shortest and smoothest paths on average. Its average path length was lower than those of the SFM (\(51.63\,\mathrm{m}\)) and MLP (\(56.63\,\mathrm{m}\)), and its path irregularity was also lower than the corresponding SFM and MLP values of \(1.93\) and \(1.83\,\mathrm{rad/m}\). The GNN and SFM achieved the same average closest pedestrian distance of \(1.82\,\mathrm{m}\), whereas the MLP achieved a marginally greater distance of \(1.83\,\mathrm{m}\). However, the GNN produced the highest minimum TTC, outperforming both the SFM (\(1.48\,\mathrm{s}\)) and MLP (\(1.27\,\mathrm{s}\)). These results indicate that explicit graph-based interaction modeling provides a better overall efficiency-smoothness-safety trade-off than either model-based social-force refinement or non-message-passing interaction pooling in previously unseen real-world environments.
\subsection{Qualitative Analysis of Interaction Patterns}
\begin{figure}
    \centering
    \includegraphics[width=1.0\columnwidth]{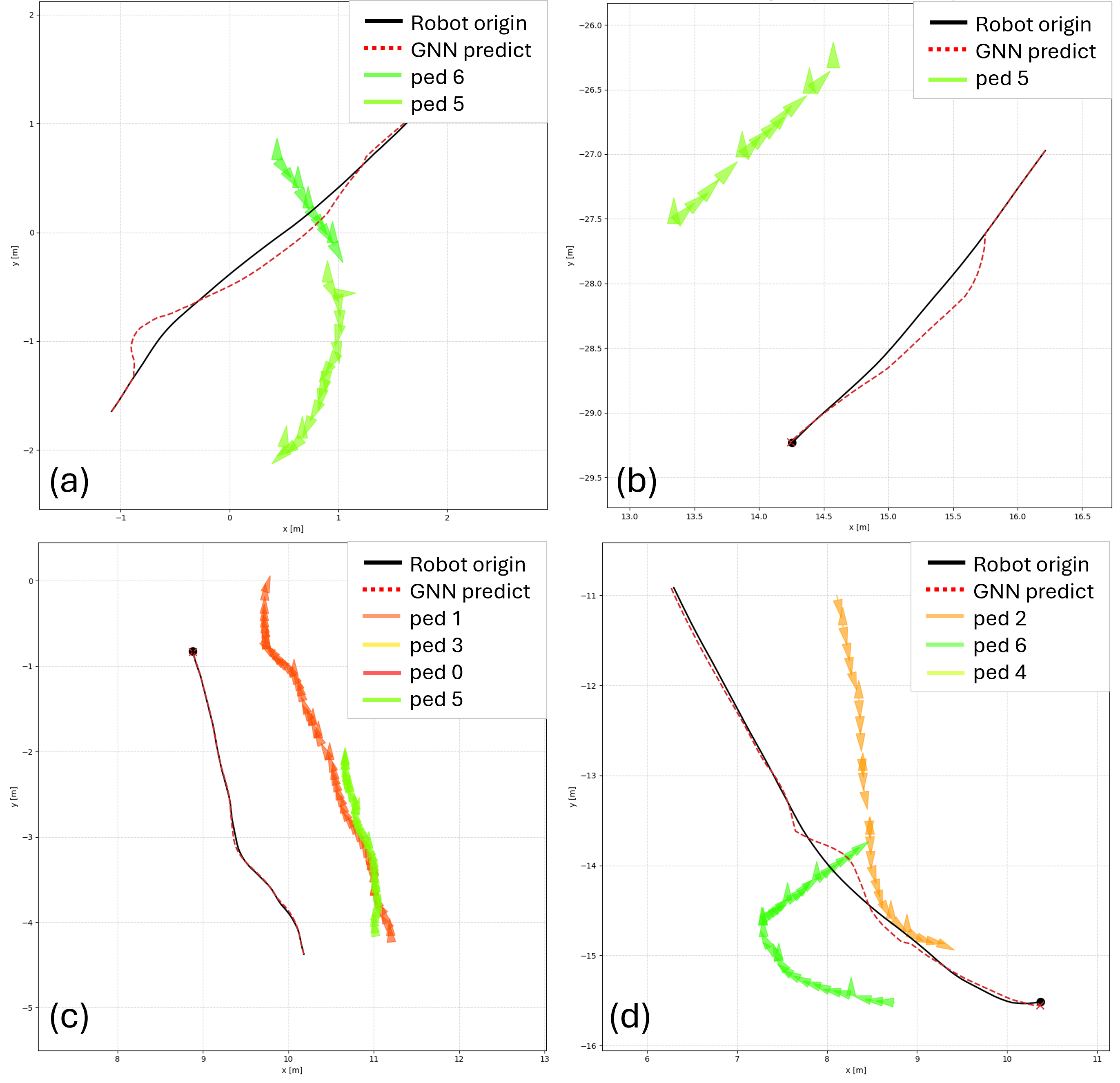}
    \caption{GNN-learned HRI Patterns: (a) crossing paths, (b) merging, (c) overtaking, and (d) group interaction.}
    \label{fg:hri_gnn}
\end{figure}

The qualitative trajectories in Fig.~\ref{fg:hri_gnn} further illustrate the different HRI patterns represented by the GNN. In the crossing path case (Fig.~\ref{fg:hri_gnn}(a)), the predicted trajectory deviated before the intersection and returned toward the original route after the pedestrians had passed, indicating an anticipatory correction. During merging (Fig.~\ref{fg:hri_gnn}(b)), the model produced a gradual lateral adjustment as the pedestrian approached the robot's direction of travel. In the overtaking case (Fig.~\ref{fg:hri_gnn}(c)), the predicted trajectory largely overlapped with the original path, indicating that the GNN did not introduce an unnecessary lateral deviation when the interaction could be handled while maintaining the nominal route. For group interaction (Fig.~\ref{fg:hri_gnn}(d)), the GNN generated a coordinated deviation in response to multiple nearby pedestrians. These examples suggest that the model produced context-dependent corrections according to pedestrian configuration and relative motion, rather than applying a uniform repulsive response in every interaction.
\subsection{Ablation study}
\subsubsection{Ablation on different prediction interval}
Table~\ref{tb:delta_t} presents the ablation study of the prediction interval \(\Delta t\) used to generate the constant-velocity residual targets. The results reveal a trade-off between trajectory efficiency and interaction safety. The \(0.5\,\mathrm{s}\) interval produced the shortest paths on both the Dynablox and bricklaying datasets and the lowest path irregularity on Dynablox. However, increasing the interval to \(1.0\,\mathrm{s}\) consistently improved the two interaction-safety metrics. On Dynablox, the closest pedestrian distance increased from \(1.77\) to \(1.82\,\mathrm{m}\), while the minimum TTC increased from \(1.42\) to \(1.77\,\mathrm{s}\). Similarly, in the bricklaying trial, the closest pedestrian distance increased from \(1.55\) to \(1.58\,\mathrm{m}\), and the minimum TTC increased from \(0.46\) to \(0.51\,\mathrm{s}\). Moreover, the \(1.0\,\mathrm{s}\) interval resulted in the lowest path irregularity in the bricklaying trial, decreasing it from \(5.05\) to \(4.39\,\mathrm{rad/m}\). These results indicate that the \(0.5\,\mathrm{s}\) interval may be too short to capture sufficiently anticipatory responses to nearby pedestrians, despite its advantage in path efficiency.

Extending \(\Delta t\) beyond \(1.0\,\mathrm{s}\) did not provide consistent additional benefits. Although the \(1.5\,\mathrm{s}\) interval yielded the greatest pedestrian clearance in the bricklaying trial and a slightly higher minimum TTC on Dynablox, it generated longer and more irregular paths on both datasets. At \(2.0\,\mathrm{s}\), the path length increased substantially to \(58.98\,\mathrm{m}\) on Dynablox and \(88.11\,\mathrm{m}\) in the bricklaying trial, while the minimum TTC decreased to \(1.13\) and \(0.40\,\mathrm{s}\), respectively. A long prediction interval may accumulate error under the constant-velocity assumption and incorporate unrelated effects of acceleration, turning, and goal-approaching behavior into the residual target. Accordingly, \(\Delta t=1.0\,\mathrm{s}\) was selected because it provided the most consistent balance among path efficiency, trajectory smoothness, pedestrian clearance, and anticipatory collision avoidance across both datasets, rather than because it achieved the best value for every individual metric.
\begin{table}[t]
 \caption{Results of ablation study on different prediction interval $\Delta t$. "Dynablox avg." reports the mean across seven sequences. Bold indicates the selected prediction interval. Standard deviations are calculated within each sequence and across sequences for "Dynablox avg."}\label{tb:delta_t}
\begin{tabular*}{\tblwidth}{@{\extracolsep{\fill}} 
  >{\raggedright\arraybackslash}p{2.0cm} 
  >{\raggedright\arraybackslash}p{1.5cm} 
  cccc @{}}
\toprule
\multirow{2}{*}{Measure} & \multirow{2}{*}{$\Delta t \quad(s)$} & \multicolumn{2}{c}{Dataset}\\
\cmidrule(l){3-4}
 & & Dynablox avg. & bricklaying\\
\midrule
\multirow{4}{*}{$\begin{array}{l} \text{Path} \\ \text{length} \\(m)\downarrow\end{array}$} & 0.5 & 48.33$\pm$26.07 & 60.18\\
 & \textbf{1.0} & 49.17$\pm$27.33 & 63.25\\
 & 1.5 & 51.10$\pm$25.64 & 65.81\\
 & 2.0 & 58.98$\pm$28.60 & 88.11\\
 \midrule
\multirow{4}{*}{$\begin{array}{l} \text{Path} \\ \text{irregularity} \\(rad/m)\downarrow\end{array}$} & 0.5 & 0.94$\pm$0.35 & 5.05\\
 & \textbf{1.0} & 1.39$\pm$0.26 & 4.39\\
 & 1.5 & 1.74$\pm$0.38 & 6.04\\
 & 2.0 & 2.11$\pm$0.26 & 5.22\\
\midrule
\multirow{4}{*}{$\begin{array}{l} \text{Closest} \\ \text{pedestrian} \\ \text{distance} \\(m)\uparrow\end{array}$} & 0.5 & 1.77$\pm$0.49 & 1.55$\pm$0.49\\
 & \textbf{1.0} & 1.82$\pm$0.49 & 1.58$\pm$0.50\\
 & 1.5 & 1.73$\pm$0.50 & 1.78$\pm$0.48\\
 & 2.0 & 1.84$\pm$0.55 & 1.59$\pm$0.53\\
\midrule
\multirow{4}{*}{$\begin{array}{l} \text{Minimum} \\ \text{TTC} \\(s)\uparrow\end{array}$} & 0.5 & 1.42$\pm$2.95 & 0.46$\pm$1.30\\
 & \textbf{1.0} & 1.77$\pm$3.27 & 0.51$\pm$1.37\\
 & 1.5 & 1.81$\pm$3.58 & 0.45$\pm$1.08\\
 & 2.0 & 1.13$\pm$2.18 & 0.40$\pm$0.92\\
\bottomrule
\end{tabular*}
\end{table}
\subsubsection{Ablation on different model variant}

Table~\ref{tb:ablation} evaluates the contributions of the target node, reflection-based data augmentation, and feature normalization. Although an explicit target node was included only when constructing the training graphs and was unavailable in the experimental trajectories, it provided useful contextual information during learning. In particular, the target node may help the model distinguish residual motion associated with goal approaching from corrections induced by nearby pedestrians. This reduced the likelihood that ordinary goal-directed deviations in the demonstrations were incorrectly learned as social responses. Consequently, removing the target node during training decreased the minimum TTC from \(1.77\) to \(1.41\,\mathrm{s}\) on Dynablox and from \(0.51\) to \(0.44\,\mathrm{s}\) in the bricklaying trial. The full model could still be applied without a target node during evaluation because its shared message-passing operations accommodate graphs containing different numbers and types of nodes.

Reflection-based data augmentation further improved generalization by exposing the model to geometrically symmetric interaction patterns that were underrepresented in the limited demonstration dataset. Without augmentation, the model generated shorter and smoother paths, but its minimum TTC decreased to \(1.41\,\mathrm{s}\) on Dynablox and \(0.36\,\mathrm{s}\) in the bricklaying trial. The particularly large reduction in the bricklaying TTC suggests that augmentation encouraged more anticipatory corrections when interaction directions and spatial configurations differed from those observed in the original demonstrations. Similarly, feature normalization prevented position, velocity, and distance variables with relatively large numerical ranges from disproportionately influencing model optimization. Removing normalization slightly shortened the paths but increased bricklaying path irregularity from \(4.39\) to \(5.51\,\mathrm{rad/m}\) and reduced the minimum TTC to \(1.44\,\mathrm{s}\) on Dynablox and \(0.46\,\mathrm{s}\) in the bricklaying trial.

The differences in closest pedestrian distance were small across all variants, indicating that these components primarily affected the timing and shape of the avoidance response rather than the final spatial clearance. The full model also did not produce the shortest or smoothest path for every dataset. Instead, it introduced moderately larger corrections while achieving the highest minimum TTC in both evaluations. Therefore, the target node, data augmentation, and normalization were retained because their combination provided the most consistent balance between trajectory quality and anticipatory interaction safety across the construction-specific and unseen real-world datasets.
\begin{table}[]
 \caption{Results of ablation study on different model variants. "Dynablox avg." reports the mean across seven sequences. Bold indicates the selected model variant. Standard deviations are calculated within each sequence and across sequences for "Dynablox avg."}\label{tb:ablation}
\begin{tabular*}{\tblwidth}{@{\extracolsep{\fill}} 
  >{\raggedright\arraybackslash}p{1.5cm} 
  >{\raggedright\arraybackslash}p{2.0cm} 
  cccc @{}}
\toprule
\multirow{2}{*}{Measure} & \multirow{2}{*}{\text{Model variant}} & \multicolumn{2}{c}{Dataset}\\
\cmidrule(l){3-4}
 & & Dynablox avg. & bricklaying\\
\midrule
\multirow{4}{*}{$\begin{array}{l} \text{Path} \\ \text{length}\\(m)\downarrow \end{array}$} & \textbf{full} & 49.17$\pm$27.33 & 63.25\\
 & \text{w/o target node} & 47.50$\pm$25.06 & 63.05\\
 & \text{w/o data aug.} & 48.99$\pm$25.06 & 62.58\\
 & \text{w/o normalization} & 47.81$\pm$25.27 & 61.95\\
 \midrule
\multirow{4}{*}{$\begin{array}{l} \text{Path} \\ \text{irregularity} \\(rad/m)\downarrow\end{array}$} & \textbf{full} & 1.39$\pm$0.26 & 4.39\\
 & \text{w/o target node} & 1.28$\pm$0.28 & 5.54\\
 & \text{w/o data aug.} & 1.17$\pm$0.30 & 5.37\\
 & \text{w/o normalization} & 1.38$\pm$0.26 & 5.51\\
\midrule
\multirow{4}{*}{$\begin{array}{l} \text{Closest} \\ \text{pedestrian} \\ \text{distance} \\(m)\uparrow\end{array}$} & \textbf{full} & 1.82$\pm$0.49 & 1.58$\pm$0.50\\
 & \text{w/o target node} & 1.82$\pm$0.49 & 1.56$\pm$0.50\\
 & \text{w/o data aug.} & 1.79$\pm$0.48 & 1.56$\pm$0.49\\
 & \text{w/o normalization} & 1.82$\pm$0.50 & 1.58$\pm$0.50\\
\midrule
\multirow{4}{*}{$\begin{array}{l} \text{Minimum} \\ \text{TTC} \\(s)\uparrow\end{array}$} & \textbf{full} & 1.77$\pm$3.27 & 0.51$\pm$1.37\\
 & \text{w/o target node} & 1.41$\pm$2.31 & 0.44$\pm$1.25\\
 & \text{w/o data aug.} & 1.41$\pm$2.59 & 0.36$\pm$0.90\\
 & \text{w/o normalization} & 1.44$\pm$2.90 & 0.46$\pm$1.17\\
\bottomrule
\end{tabular*}
\end{table}
\section{Discussion}
Regarding the first research problem, the results demonstrate that constant-velocity residuals can serve as effective automatically generated targets for learning interaction-aware path corrections from unlabeled demonstrations. Compared with the original trajectories, the GNN generally improved minimum TTC, indicating that the learned residuals captured anticipatory responses to potential conflicts rather than merely reproducing nominal motion. The prediction-horizon ablation further showed that the quality of these targets depends on their temporal scale. A \(0.5\,\mathrm{s}\) horizon favored shorter paths but provided weaker anticipatory responses, whereas horizons longer than \(1.0\,\mathrm{s}\) produced increasingly long and irregular trajectories because errors accumulated under the constant-velocity assumption. The selected \(1.0\,\mathrm{s}\) horizon achieved the most consistent balance across both datasets. These findings suggest that meaningful interaction behavior can be extracted without manually annotating socially preferred paths, although the residuals may also contain effects arising from acceleration, turning, and goal-directed motion.

As for the second research problem, the comparison with the SFM and MLP indicates that explicit graph-based interaction modeling provides a favorable balance among efficiency, smoothness, and anticipatory collision avoidance. The GNN produced shorter and smoother refined trajectories than both baselines while achieving the highest minimum TTC across the Dynablox sequences and bricklaying trial. Its relatively small improvement in minimum pedestrian distance, accompanied by a larger improvement in TTC, suggests that the model primarily learned to adjust the timing and direction of its response instead of simply maximizing spatial clearance. The ablation results also show that the target node helps distinguish goal-directed motion from interaction-induced deviations during training, while reflection augmentation and normalization improve transfer to unfamiliar configurations. Although the target node is omitted during deployment to avoid dependence on planner-provided goal information, the spatial residual output can still be applied to paths from different planners, supporting a planner-agnostic correction interface rather than a planner-specific control policy.

In terms of the third research problem, the Dynablox-EKF-GNN pipeline demonstrates that frame-wise LiDAR detections can be converted into continuous dynamic-object states and used for real-time interaction-aware path refinement. The EKF associates consecutive detections and estimates object positions and velocities, providing the temporally consistent node and edge features required by the GNN. Tracks are created, updated, or removed according to spatial and temporal association thresholds, enabling the system to follow continuously moving objects while accommodating objects entering or leaving the sensing range. Because graph construction and GNN inference operate directly on these tracked states, the framework does not require camera observations, appearance-based identities, or offline trajectory processing. Its application to both the construction-specific bricklaying trial and the independent Dynablox sequences indicates that the pipeline can operate across different environments and interaction patterns. Overall, the findings support the feasibility of a real-time, LiDAR-based perception-to-refinement pipeline, with the GNN consuming dynamically constructed graphs during robot operation.
\section{Conclusion}
\subsection{Summary}
This study presents a real-time, LiDAR-based framework for learning interaction-aware path corrections from unlabeled human demonstrations. Constant-velocity residuals were used as automatically generated learning targets, while a GNN modeled interactions among the robot, pedestrians, and target during training. An EKF-based tracking module converted frame-wise LiDAR detections into continuous position and velocity estimates for real-time graph construction. Evaluations on the construction-specific bricklaying trial and the independent Dynablox dataset showed that the proposed method achieved a favorable balance among path efficiency, smoothness, pedestrian clearance, and minimum TTC compared with the original paths, SFM, and MLP baseline. The ablation studies further supported the selection of a \(1.0\,\mathrm{s}\) prediction interval and demonstrated the benefits of the target node, reflection augmentation, and feature normalization.
\subsection{Limitations and future work}
Despite the success, this study still poses several limitations. First, the real-world validation was conducted in a limited range of construction scenarios using a single robot platform. Therefore, the findings may not fully generalize to other tasks, layouts, interaction patterns, or robot configurations. Second, the constant-velocity assumption used to generate learning targets may not accurately represent acceleration or sudden directional changes. Third, the evaluation relies primarily on objective metrics, including pedestrian distance and time-to-collision, which do not capture workers' perceived comfort, motion naturalness, or social acceptability. These limitations constrain the generalizability and interpretation of the reported interaction-aware navigation performance.

Future work will first evaluate the framework across more diverse construction tasks, environmental layouts, interaction patterns, and robot platforms to establish its generalizability. Second, more expressive motion-prediction and target-generation methods will be investigated to address the limitations of the constant-velocity assumption and better separate socially motivated responses from ordinary acceleration, turning, and goal-directed behavior. Finally, human-centered experiments will be conducted to assess workers' perceived comfort, motion naturalness, and social acceptability. Combining these subjective evaluations with objective measures such as pedestrian distance and TTC would provide a more comprehensive assessment of the social quality of the generated trajectories.
\printcredits
\section*{Declaration of competing interest}
The authors declare that they have no known competing financial interests or personal relationships that could have appeared to influence the work reported in this paper.
\section*{Acknowledgments}
This material is based upon work supported by the National Science Foundation under Grant No. 2222723. Any opinions, findings, and conclusions or recommendations expressed in this material are those of the authors and do not necessarily reflect the views of the National Science Foundation.
\section*{Data Availability}
The authors do not have permission to share data.
\bibliographystyle{cas-model2-names}
\bibliography{cas-refs}

@article{KIM2024105455,
title = {Understanding human-robot proxemic norms in construction: How do humans navigate around robots?},
journal = {Automation in Construction},
volume = {164},
pages = {105455},
year = {2024},
issn = {0926-5805},
doi = {10.1016/j.autcon.2024.105455},
url = {https://www.sciencedirect.com/science/article/pii/S0926580524001912},
author = {YeSeul Kim and Seongyong Kim and Yilong Chen and HyunJin Yang and Seungwoo Kim and Sehoon Ha and Matthew Gombolay and Yonghan Ahn and Yong Kwon Cho},
}

@article{CHARALAMPOUS201785,
title = {Recent trends in social aware robot navigation: A survey},
journal = {Robotics and Autonomous Systems},
volume = {93},
pages = {85-104},
year = {2017},
issn = {0921-8890},
doi = {10.1016/j.robot.2017.03.002},
url = {https://www.sciencedirect.com/science/article/pii/S0921889016302287},
author = {Konstantinos Charalampous and Ioannis Kostavelis and Antonios Gasteratos},
}

@ARTICLE{Gao2022,
AUTHOR={Gao, Yuxiang  and Huang, Chien-Ming }, 
TITLE={Evaluation of Socially-Aware Robot Navigation}, 
JOURNAL={Frontiers in Robotics and AI}, 
VOLUME={Volume 8 - 2021},
YEAR={2022},
URL={https://www.frontiersin.org/journals/robotics-and-ai/articles/10.3389/frobt.2021.721317},
DOI={10.3389/frobt.2021.721317},
ISSN={2296-9144},
}

@article{kim2025understanding,
  title={Understanding the effects of humanlike robot motions on unfocused human--robot interaction},
  author={Kim, YeSeul and Chen, Yilong and Gombolay, Mathew and Cho, Yong K},
  journal={Advanced Engineering Informatics},
  volume={68},
  pages={103751},
  year={2025},
  publisher={Elsevier}
}

@article{Chen2026AnOF,
  title={An Observation Feature Study of Robot Imitation Learning for Autonomous Social Navigation on Construction Sites},
  author={Yilong Chen and Yu Du and Xiaoke Wang and Yong K. Cho},
  journal={Computing in Civil Engineering 2025},
  year={2026},
  url={https://api.semanticscholar.org/CorpusID:285166161}
}

@article{Chen2026Enhance,
  title={Enhancing Socially Aware Construction Robot Navigation through LiDAR-Camera Sensor Fusion and Dynamic Object Tracking},
  author={Yilong Chen and Yong K. Cho},
  journal={Journal of Computing in Civil Engineering},
  year={2026},
  DOI={10.1061/JCCEE5.CPENG-7690},
}

@article{JIA2023104984,
title = {Graph neural networks for construction applications},
journal = {Automation in Construction},
volume = {154},
pages = {104984},
year = {2023},
issn = {0926-5805},
doi = {10.1016/j.autcon.2023.104984},
url = {https://www.sciencedirect.com/science/article/pii/S0926580523002443},
author = {Yilong Jia and Jun Wang and Wenchi Shou and M. Reza Hosseini and Yu Bai},
}

@INPROCEEDINGS{GoriGNN,
  author={Gori, M. and Monfardini, G. and Scarselli, F.},
  booktitle={Proceedings. 2005 IEEE International Joint Conference on Neural Networks, 2005.}, 
  title={A new model for learning in graph domains}, 
  year={2005},
  volume={2},
  number={},
  pages={729-734 vol. 2},
  doi={10.1109/IJCNN.2005.1555942}}

@ARTICLE{MicheliMP,
  author={Micheli, Alessio},
  journal={IEEE Transactions on Neural Networks}, 
  title={Neural Network for Graphs: A Contextual Constructive Approach}, 
  year={2009},
  volume={20},
  number={3},
  pages={498-511},
  doi={10.1109/TNN.2008.2010350}}

@inproceedings{GilmerMP,
author = {Gilmer, Justin and Schoenholz, Samuel S. and Riley, Patrick F. and Vinyals, Oriol and Dahl, George E.},
title = {Neural message passing for Quantum chemistry},
year = {2017},
publisher = {JMLR.org},
booktitle = {Proceedings of the 34th International Conference on Machine Learning - Volume 70},
pages = {1263–1272},
numpages = {10},
location = {Sydney, NSW, Australia},
series = {ICML'17}
}

@InProceedings{MansoSN,
author="Manso, Luis J.
and Jorvekar, Ronit R.
and Faria, Diego R.
and Bustos, Pablo
and Bachiller, Pilar",
editor="Bergasa, Luis M.
and Oca{\~{n}}a, Manuel
and Barea, Rafael
and L{\'o}pez-Guill{\'e}n, Elena
and Revenga, Pedro",
title="Graph Neural Networks for Human-Aware Social Navigation",
booktitle="Advances in Physical Agents II",
year="2021",
publisher="Springer International Publishing",
address="Cham",
pages="167--179",
isbn="978-3-030-62579-5"
}

@article{bachiller2022graph,
  title={A graph neural network to model disruption in human-aware robot navigation},
  author={Bachiller, Pilar and Rodriguez-Criado, Daniel and Jorvekar, Ronit R and Bustos, Pablo and Faria, Diego R and Manso, Luis J},
  journal={Multimedia tools and applications},
  volume={81},
  number={3},
  pages={3277--3295},
  year={2022},
  publisher={Springer}
}

@INPROCEEDINGS{ZhaoSim2Real,
  author={Zhao, Wenshuai and Queralta, Jorge Peña and Westerlund, Tomi},
  booktitle={2020 IEEE Symposium Series on Computational Intelligence (SSCI)}, 
  title={Sim-to-Real Transfer in Deep Reinforcement Learning for Robotics: a Survey}, 
  year={2020},
  volume={},
  number={},
  pages={737-744},
  doi={10.1109/SSCI47803.2020.9308468}}

@ARTICLE{WuSim2Real,
  author={Wu, Jingda and Zhou, Yanxin and Yang, Haohan and Huang, Zhiyu and Lv, Chen},
  journal={IEEE Transactions on Pattern Analysis and Machine Intelligence}, 
  title={Human-Guided Reinforcement Learning With Sim-to-Real Transfer for Autonomous Navigation}, 
  year={2023},
  volume={45},
  number={12},
  pages={14745-14759},
  doi={10.1109/TPAMI.2023.3314762}}

@article{GervetSR,
  author  = {Theophile Gervet  and Soumith Chintala  and Dhruv Batra  and Jitendra Malik  and Devendra Singh Chaplot },
  title   = {Navigating to objects in the real world},
  journal = {Science Robotics},
  volume  = {8},
  number  = {79},
  pages   = {eadf6991},
  year    = {2023},
  doi     = {10.1126/scirobotics.adf6991},
  url     = {https://www.science.org/doi/abs/10.1126/scirobotics.adf6991},
  eprint  = {https://www.science.org/doi/pdf/10.1126/scirobotics.adf6991}
}

@misc{xu2026navrlsystemlevelframeworkimproving,
      title={NavRL++: A System-Level Framework for Improving Sim-to-Real Transfer in Reinforcement Learning-Based Robot Navigation}, 
      author={Zhefan Xu and Hanyu Jin and Kenji Shimada},
      year={2026},
      eprint={2605.15559},
      archivePrefix={arXiv},
      primaryClass={cs.RO},
      url={https://arxiv.org/abs/2605.15559}, 
}

@book{slam-handbook,
  title = {{SLAM Handbook. } From Localization and Mapping to Spatial Intelligence},
  editor = {Luca Carlone and Ayoung Kim and Timothy Barfoot and Daniel Cremers and Frank Dellaert},
  publisher = {Cambridge University Press},
  year = {2026}
}

@article{Schmid_2023,
   title={Dynablox: Real-Time Detection of Diverse Dynamic Objects in Complex Environments},
   volume={8},
   ISSN={2377-3774},
   url={http://dx.doi.org/10.1109/LRA.2023.3305239},
   DOI={10.1109/lra.2023.3305239},
   number={10},
   journal={IEEE Robotics and Automation Letters},
   publisher={Institute of Electrical and Electronics Engineers (IEEE)},
   author={Schmid, Lukas and Andersson, Olov and Sulser, Aurelio and Pfreundschuh, Patrick and Siegwart, Roland},
   year={2023},
   month=Oct, pages={6259–6266} }

@inproceedings{chen2025online,
  author    = {Chen, Y. and Kim, Y. S. and Yarovoi, A. and Kim, S. and Cho, Y. K.},
  title     = {Online Dynamic Object Detection on Construction Sites Using SLAM and Occupancy Grid},
  booktitle = {Computing in Civil Engineering},
  pages     = {556--563},
  year      = {2025},
  publisher = {American Society of Civil Engineers},
  doi       = {10.1061/9780784486115.069},
  url       = {https://ascelibrary.org/doi/10.1061/9780784486115.069}
}

@misc{chen2026icra,
      title={Dynamic Object Detection and Tracking in Construction: A Fisheye Camera and LiDAR Sensor Fusion Model}, 
      author={Yilong Chen and Huili Huang and Yong K. Cho},
      year={2026},
      eprint={2607.06896},
      archivePrefix={arXiv},
      primaryClass={cs.RO},
      url={https://arxiv.org/abs/2607.06896}, 
}

@inproceedings{kim2024much,
  title={How much distance should robots keep from other workers at construction jobsites?},
  author={Kim, YeSeul and Chen, Yilong and Kim, Seongyong and Cho, Yong K},
  booktitle={Construction Research Congress 2024},
  pages={893--902}
}

@article{Chen2026Social,
  title={Socially Aware Construction Robot Navigation with a GNN-Enhanced DWA Planner},
  author={Yilong Chen and Yong K. Cho},
  journal={Computing in Civil Engineering 2026},
  year={2026}}

@misc{wang2023graphneuralnetworklocal,
      title={Graph Neural Network with Local Frame for Molecular Potential Energy Surface}, 
      author={Xiyuan Wang and Muhan Zhang},
      year={2023},
      eprint={2208.00716},
      archivePrefix={arXiv},
      primaryClass={cs.LG},
      url={https://arxiv.org/abs/2208.00716}, 
}

@article{Helbing_1995,
   title={Social force model for pedestrian dynamics},
   volume={51},
   ISSN={1095-3787},
   url={http://dx.doi.org/10.1103/PhysRevE.51.4282},
   DOI={10.1103/physreve.51.4282},
   number={5},
   journal={Physical Review E},
   publisher={American Physical Society (APS)},
   author={Helbing, Dirk and Molnár, Péter},
   year={1995},
   month=May, pages={4282–4286} }

@INPROCEEDINGS{GuzziHuman,
  author={Guzzi, Jérôme and Giusti, Alessandro and Gambardella, Luca M. and Theraulaz, Guy and Di Caro, Gianni A.},
  booktitle={2013 IEEE International Conference on Robotics and Automation}, 
  title={Human-friendly robot navigation in dynamic environments}, 
  year={2013},
  volume={},
  number={},
  pages={423-430},
  doi={10.1109/ICRA.2013.6630610}}

@article{BiswasSocNav,
author = {Biswas, Abhijat and Wang, Allan and Silvera, Gustavo and Steinfeld, Aaron and Admoni, Henny},
title = {SocNavBench: A Grounded Simulation Testing Framework for Evaluating Social Navigation},
year = {2022},
issue_date = {September 2022},
publisher = {Association for Computing Machinery},
address = {New York, NY, USA},
volume = {11},
number = {3},
url = {https://doi.org/10.1145/3476413},
doi = {10.1145/3476413},
journal = {J. Hum.-Robot Interact.},
month = jul,
articleno = {26},
numpages = {24}
}

@article{Francis2025,
author = {Francis, Anthony and P\'{e}rez-D’Arpino, Claudia and Li, Chengshu and Xia, Fei and Alahi, Alexandre and Alami, Rachid and Bera, Aniket and Biswas, Abhijat and Biswas, Joydeep and Chandra, Rohan and Chiang, Hao-Tien Lewis and Everett, Michael and Ha, Sehoon and Hart, Justin and How, Jonathan P. and Karnan, Haresh and Lee, Tsang-Wei Edward and Manso, Luis J. and Mirsky, Reuth and Pirk, S\"{o}ren and Singamaneni, Phani Teja and Stone, Peter and Taylor, Ada V. and Trautman, Peter and Tsoi, Nathan and V\'{a}zquez, Marynel and Xiao, Xuesu and Xu, Peng and Yokoyama, Naoki and Toshev, Alexander and Mart\'{\i}n-Mart\'{\i}n, Roberto},
title = {Principles and Guidelines for Evaluating Social Robot Navigation Algorithms},
year = {2025},
issue_date = {June 2025},
publisher = {Association for Computing Machinery},
address = {New York, NY, USA},
volume = {14},
number = {2},
url = {https://doi.org/10.1145/3700599},
doi = {10.1145/3700599},
journal = {J. Hum.-Robot Interact.},
month = feb,
articleno = {34},
numpages = {65}
}

@article{fox1997dynamic,
  title={The dynamic window approach to collision avoidance},
  author={Fox, Dieter and Burgard, Wolfram and Thrun, Sebastian},
  journal={IEEE robotics \& automation magazine},
  volume={4},
  number={1},
  pages={23--33},
  year={1997},
  publisher={IEEE}
}

@article{rosmann2017integrated,
  title={Integrated online trajectory planning and optimization in distinctive topologies},
  author={R{\"o}smann, Christoph and Hoffmann, Frank and Bertram, Torsten},
  journal={Robotics and Autonomous Systems},
  volume={88},
  pages={142--153},
  year={2017},
  publisher={Elsevier}
}

\end{document}